\documentclass[sigconf]{acmart}
\usepackage{tabu}
\usepackage{arydshln}
\usepackage{subfigure}
\usepackage{booktabs}
\usepackage{multirow}
\usepackage{tabularx}
\usepackage{graphicx}
\usepackage{comment}
\usepackage{amsmath}
\usepackage{xcolor}
\usepackage{wasysym}
\usepackage[boxed, noend]{algorithm2e}
\usepackage{lscape}
\usepackage{rotating}
\usepackage[colorinlistoftodos]{todonotes}
\usepackage{longtable} 
\usepackage{pdflscape} 
\usepackage{adjustbox, array, placeins}
\DeclareMathOperator*{\argmax}{arg\,max}
\AtBeginDocument{%
  \providecommand\BibTeX{{%
    \normalfont B\kern-0.5em{\scshape i\kern-0.25em b}\kern-0.8em\TeX}}}

\newcommand\blfootnote[1]{%
  \begingroup
  \renewcommand\thefootnote{}\footnote{#1}%
  \addtocounter{footnote}{-1}%
  \endgroup
}

\copyrightyear{2020}
\acmYear{2020}
\setcopyright{acmlicensed}
\acmConference[ACM CHIL '20]{ACM Conference on Health, Inference, and Learning}{April 2--4, 2020}{Toronto, ON, Canada}
\acmBooktitle{ACM Conference on Health, Inference, and Learning (ACM CHIL '20), April 2--4, 2020, Toronto, ON, Canada}
\acmPrice{15.00}
\acmDOI{10.1145/3368555.3384448}
\acmISBN{978-1-4503-7046-2/20/04}

\begin{document}

\title{Hurtful Words: Quantifying Biases in Clinical Contextual Word Embeddings}


\author{Haoran Zhang*}
\email{haoran@cs.toronto.edu}
\affiliation{%
\institution{University of Toronto}
}
\affiliation{%
\institution{Vector Institute}
}

\author{Amy X. Lu*}
\email{amyxlu@cs.toronto.edu}
\affiliation{%
\institution{University of Toronto}
}
\affiliation{%
\institution{Vector Institute}
}

\author{Mohamed Abdalla}
\email{msa@cs.toronto.edu}
\affiliation{%
\institution{University of Toronto}
}
\affiliation{%
\institution{Vector Institute}
}

\author{Matthew McDermott}
\email{mmd@mit.edu}
\affiliation{%
\institution{Massachusetts Institute of Technology}
}
\author{Marzyeh Ghassemi}
\email{marzyeh@cs.toronto.edu}
\affiliation{%
\institution{University of Toronto}
}
\affiliation{%
\institution{Vector Institute}
}

\renewcommand{\shortauthors}{Zhang and Lu, et al.}
\begin{abstract}
In this work, we examine the extent to which embeddings may encode marginalized populations differently, and how this may lead to a perpetuation of biases and worsened performance on clinical tasks.
We pretrain deep embedding models (BERT) on medical notes from the MIMIC-III hospital dataset, and quantify potential disparities using two approaches. First, we identify dangerous latent relationships that are captured by the contextual word embeddings using a fill-in-the-blank method with text from real clinical notes and a log probability bias score quantification. Second, we evaluate performance gaps across different definitions of fairness on over 50 downstream clinical prediction tasks that include detection of acute and chronic conditions. 
We find that classifiers trained from BERT representations exhibit statistically significant differences in performance, often favoring the majority group with regards to gender, language, ethnicity, and insurance status. 
Finally, we explore shortcomings of using adversarial debiasing to obfuscate subgroup information in contextual word embeddings, and recommend best practices for such deep embedding models in clinical settings.

\end{abstract}



\maketitle
\blfootnote{* Equal Contribution} 
\section{Introduction}
As deployable machine learning systems begin to play a role in society, there is a question of how best to benefit from technology while maintaining or improving fairness. There are documented effects of bias propagated by real-world machine learning systems, from gender biases in job classification \cite{de2019bias}, to decreased prediction accuracies on dark-skinned females due to sample imbalances \cite{buolamwini2018gender}, and to overprescribing policing resources to historically over-policed neighbourhoods \cite{lum2016to}.

\begin{figure}[h!]
    \centering
    \includegraphics[width=0.47\textwidth]{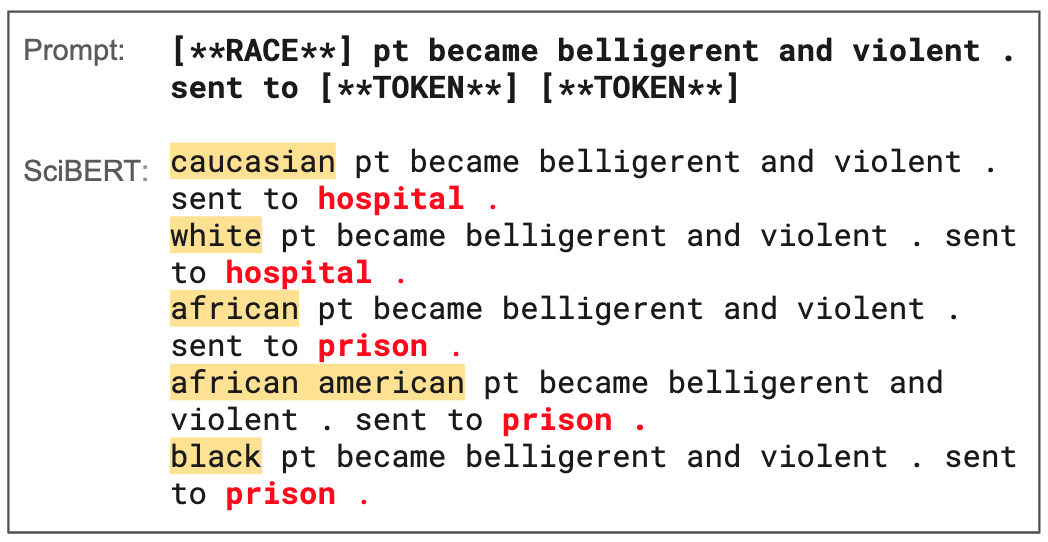}
    \caption{When prompted to generate course of action in a fill-in-the-blank task, SciBERT~\cite{Beltagy2019} generates different results for different races. Templates are adapted from real clinical notes in the MIMIC-III database \cite{Johnson2016}, where the shorthand ``pt" abbreviates ``patient". More detailed methods can be found in Appendix \ref{sec:fill_in_blank_procedure}.}
    \label{fig:fillinblanks}
\end{figure}

In clinical applications, machine learning has the potential to improve patient outcomes, cut costs, and reduce physician burnout \cite{Ghassemi2018,topol2019high,seyyed2020chexclusion}. Many machine learning algorithms are already in use for healthcare applications: e.g., IBM Watson has been used to support diagnoses and planning for oncology patients \cite{ibmwatson} and Novartis has developed tools to monitor and predict trial enrollment, costs and quality \cite{novartis}. Importantly, as algorithms enter healthcare systems, it is important not to exacerbate the treatment disparities amongst existing subgroups~\cite{glance2013trends,meghani2012time,schwartz2014racial}. Machine learning models trained on healthcare data may exhibit such biases \cite{chen2019can,chen2018my, Yu238}, and it is important to take measures that minimize the impact of bias on predictive systems \cite{vayena2018machine, nelson2019bias}.

In this work, we characterize the biases that can be operationalized in clinical prediction tasks by training state-of-the-art word embedding models on unstructured clinical notes. We focus on word embedding algorithms that transform text into dense numeric vectors \cite{suresh2017clinical}. While these techniques have high predictive ability, they can capture relationships between words that reflect biases, e.g., along gender \cite{bolukbasi2016man} or ethnic identity \cite{garg2018word}. While such biases exist within both traditional non-contextual word embeddings~\cite{bolukbasi2016man,zhao2018learning} and contextual word embeddings~\cite{zhao2019gender,kurita2019measuring, tan2019assessing}, biases in pretrained contextual embedding models can additionally encode historical biases in the training corpora, class imbalance in datasets, and data quality differences \cite{rajkomar2018ensuring, zou2018ai}. This is especially concerning with the growth of large \emph{pretrained} contextual embeddings models, such as variants of the BERT architecture \cite{devlin2019bert,liu2019roberta,yang2019xlnet,peters2018deep,radford2019language}.

In order to assess the impact of biases of pretrained systems, we train a BERT model initialized from SciBERT, a public BERT model pretrained on scientific text~\cite{Beltagy2019}, on the clinical notes found in the MIMIC-III database \cite{Johnson2016}, generating a baseline clinical BERT model. As a motivating example, in Figure \ref{fig:fillinblanks}, we present a sample medical word completion task using SciBERT to generate medical context given patient race. As shown, the modification of race generates a worse course of action for African American patients, which could lead to further discrimination in the healthcare system. We target three specific investigations of whether the baseline clinical BERT model is ``fair'' along four categories of protected attributes: gender, language spoken, ethnicity, and insurance type. In order to quantify how ``fair'' the model is, we must turn to a definition of fairness from the literature. There are many technical definitions of fairness, and specific arguments for or against using each in healthcare settings~\cite{hardt2016equality, Pfohl2019, Kleinberg2017}. In this work, we examine the demographic parity, equality of opportunity for the
positive class, and equality of opportunity for the negative class.\footnote{Here, ``positive" and ``negative" class refers to the label during binary classification.} 

\begin{figure*}[ht!]
    \centering
    \includegraphics[width=0.8\textwidth]{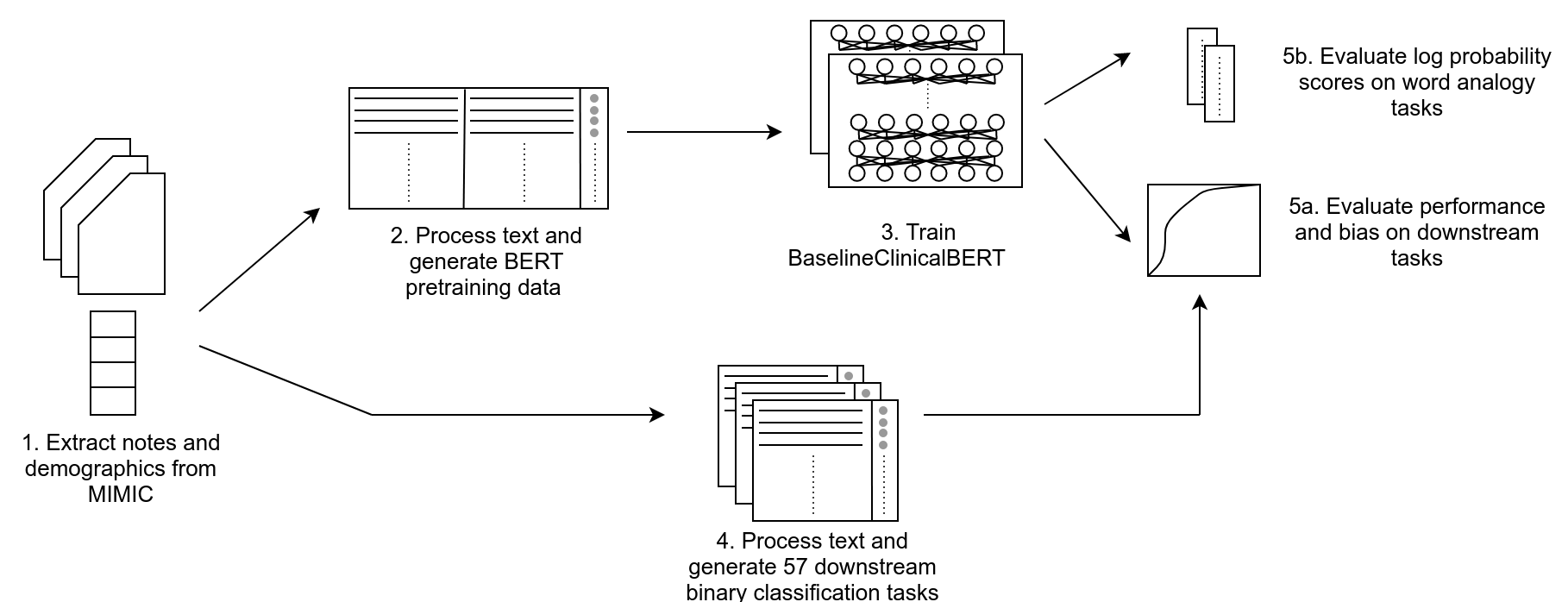}
    \caption{Process flow for extracting and preparing data, model training, and model evaluation. We extract notes from the MIMIC-III database, and then pretrain a BERT model on this data. Subsequently, we construct a cohort from MIMIC-III and create 57 binary classification problems, which we use alongside the log probability score method to evaluate bias in the pretrained model. }
    \label{fig:workflow}
\end{figure*}

First, we demonstrate that there are significant differences in the log probability bias scores~\cite{kurita2019measuring} of clinical text for different genders. These scores examine the probability of filling in the gender demographics given medical context. 
Second, we show that our baseline clinical BERT exhibits predictive task performance gaps~\cite{chen2019can} across a majority of 57 downstream clinical prediction tasks. While some variance in performance is expected, we find statistically significant gaps in parity, recall and specificity in all protected attributes.  
Finally, we attempt to correct for the baseline clinical BERT's performance disparities using adversarial debiasing during pretraining~\cite{Beutel2017,Madras2018,zhang2018mitigating,wang2018adversarial}, where a discriminator model forces learned embeddings to be minimally predictive of protected subgroup information. We find that such ``de-biasing'' does not greatly reduce
the number of statistically significant gaps, indicating much need for more research in this area.
Full details of our workflow is shown in Figure~\ref{fig:workflow}.

A summary of our specific contributions are as follows:
\begin{itemize}
    \item We demonstrate that contextual embedding models (BERT specifically) trained on clinical notes exhibits differences in performance for different genders, ethnicities, language speakers, and insurance statuses. To the best of our knowledge, we are the first to do so.
    \item We apply adversarial pretraining debiasing, and find that, consistent with existing work \cite{elazar2018adversarial}, group disparities can remain in the ``debiased'' word embeddings during post-hoc classification. 
    \item We publicly release our pretrained BERT model and code to help accelerate this area of research.\footnote{Pretrained models and code: https://github.com/MLforHealth/HurtfulWords} 
\end{itemize}

The remainder of our paper is structured as follows: In Section 2, we describe previous work on contextual word embeddings and fairness of word embeddings. In Section 3, we outline the data and predictive tasks we use, as well as our clinical BERT pretraining procedure. In Section 4, we describe the fairness definitions we use, along with methods for downstream finetuning, evaluating log probability scores, and adversarial debiasing. In Section 5, we present our results, and we discuss their implications in Section 6. We discuss the limitations of our work in Section 7, and make concluding remarks in Section 8.

\section{Background and Related Work}


\subsection{Contextual Embeddings}
Word embeddings algorithms are methods for numerically representing human text as dense high-dimensional vectors which are amenable to further computational methods~\cite{lecun2015deep}. There are a wide variety of popular word embedding algorithms, and although all word embeddings use context in creating word representations, as they are based on the distributional hypothesis~\cite{Sahlgren:2008}, they can largely be classified as either contextual or non-contextual.

Non-contextual word embeddings such as Word2Vec~\cite{NIPS2013_5021} and GloVe~\cite{pennington_2014}, once trained, do not change the representation of a word given its surrounding context. Contextual word embeddings, such as ELMo~\cite{peters2018deep} and BERT~\cite{devlin2019bert}, change the representation of a word given its surrounding context. 


Contextual word embeddings are often pretrained on a large dataset through self-supervised tasks, then released for fine-tuned use in downstream tasks. This pretraining can be domain specific, such as in the various clinical-text specific BERT models which have been released~\cite{Si2019,Alsentzer2019,Huang2019}. This pretraining task can be another source in which bias present in training text can be hard-coded into a word embedding model.

\subsection{Fairness of Word Embeddings}



Word embedding models trained on large corpora have been shown to capture societal biases in addition to the intended semantic and syntactic properties of natural language, ranging from gender~\cite{bolukbasi2016man} to race and ethnicity~\cite{garg2018word}. Analyzing the biases encoded by an embedding model is not straightforward, as simple tests can often hide existing bias if not robustly tested for~\cite{gonen2019lipstick}.

To combat this, a growing community of researchers is actively working to address and remove biases from (i.e. debiasing) word embeddings. While individual works study how contextual word embeddings capture biases~\cite{basta2019evaluating,kurita2019measuring,tan2019assessing}, to date the creation of debiasing methods has been limited to non-contextual word embeddings models (e.g. GLoVe~\cite{pennington_2014}, Word2Vec~\cite{NIPS2013_5021}). 

\subsection{Pretrained Clinical Embeddings}

Several BERT models pretrained on MIMIC notes are publicly available~\cite{Alsentzer2019, Huang2019, Si2019}. However, to address several limitations, we choose to train our own clinical BERT model in this work. First, existing models are initialized from BioBERT~\cite{lee2019biobert} or BERT\textsubscript{BASE}~\cite{devlin2019bert}, though SciBERT~\cite{Beltagy2019} outperforms BioBERT on a number of downstream tasks. Secondly, existing models do not satisfactorily encode the personal health identifiers (PHI) within the notes (e.g., \textit{[**2126-9-19**]}), either leaving them as is, or removing them altogether. Thirdly, no clinical BERT model uses whole-word masking, a recent amendment to BERT pretraining which has been found to improve performance. Our publicly released model addresses these three shortcomings.



\section{Data and Pre-Training}
\subsection{Data Source}
The Multiparameter Intelligence Monitoring in Intensive Care (MIMIC-III) dataset consists of electronic health records for 38,597 adult patients and 7,870 neonates admitted to the intensive care unit of the Beth Israel Deconess Medical Center between 2001 and 2012~\cite{Johnson2016}. The dataset contains approximately 2 million clinical notes of varying types (e.g. discharge summaries, nursing notes, radiology reports), as well as patient demographic information such as ethnicity, language spoken, and insurance status.

\subsection{Clinical Prediction Tasks}
\begin{table}[h!]
\begin{adjustbox}{max width=\columnwidth}
\begin{tabular}{|>{\hangindent=2em}p{5cm}|c|c|}
\hline
\multicolumn{1}{|c|}{\textbf{Task}}                                            & \textbf{Shorthand} & \textbf{Prevalence} \\ \hline
In-hospital Mortality                                                          & Inhosp Mort        & 13.21\%                     \\ \hdashline
Acute and unspecified renal failure                                            & Acute Renal        & 20.10\%                     \\
Acute cerebrovascular disease                                                  & Cerebrovascular    & 7.08\%                      \\
Acute myocardial infarction                                                    & Myocardial         & 11.43\%                     \\
Cardiac dysrhythmias                                                           & Dysrhythmias       & 31.56\%                     \\
Chronic kidney disease                                                         & Chronic Kidney     & 11.42\%                     \\
Chronic obstructive pulmonary disease and bronchiectasis                       & COPD               & 12.91\%                     \\
Complications of surgical procedures or medical care                           & Comp. Surgical     & 20.40\%                     \\
Conduction disorders                                                           & Conduction         & 6.76\%                      \\
Congestive heart failure; nonhypertensive                                      & Heart Failure      & 27.73\%                     \\
Coronary atherosclerosis and other heart disease                               & Atherosclerosis    & 32.66\%                     \\
Diabetes mellitus with complications                                           & Diabetes Comp      & 9.09\%                      \\
Diabetes mellitus without complication                                         & Diabetes No Comp   & 18.88\%                     \\
Disorders of lipid metabolism                                                  & Lipid Metabolism   & 25.87\%                     \\
Essential hypertension                                                         & Hypertension       & 41.07\%                     \\
Fluid and electrolyte disorders                                               & Fluid Disorder     & 23.67\%                     \\
Gastrointestinal hemorrhage                                                    & GI Hemorrhage      & 7.31\%                      \\
Hypertension with complications and secondary hypertension                     & Hypertension Comp  & 11.99\%                     \\
Other liver diseases                                                           & Other Liver        & 7.76\%                      \\
Other lower respiratory disease                                                & Lower Resp         & 4.11\%                      \\
Other upper respiratory disease                                                & Upper Resp         & 3.71\%                      \\
Pleurisy; pneumothorax; pulmonary collapse                                     & Pleurisy           & 8.50\%                      \\
Pneumonia (not caused by tuberculosis or sexually transmitted disease)         & Pneumonia          & 13.96\%                     \\
Respiratory failure; insufficiency; arrest (adult)                             & Resp Failure       & 17.53\%                     \\
Septicemia (except in labor)                                                   & Septicemia         & 14.03\%                     \\
Shock                                                                          & Shock              & 7.21\%                      \\ \hdashline
Any Chronic                                                                    & Chronic            & 78.63\%                     \\
Any Acute                                                                      & Acute              & 78.56\%                     \\
Any Disease                                                                    & Disease            & 92.68\%                     \\ \hline
\end{tabular}
\end{adjustbox}
\caption{\label{tab:shorthand} Downstream clinical tasks performed, their associated abbreviations, and their prevalence. In-hospital mortality is a clinical outcome prediction task. The next set of tasks are derived from ICD-9 billing codes (prevalence shown are for the \textit{phenotype using all notes} cohort). The final three tasks are derived from logical ORs on subsets of the previous set of tasks. }
\vspace{-1.5em}
\end{table}

To evaluate the performance of our baseline and debiased BERT models, we use the following downstream binary classification clinical tasks.

\paragraph{In-hospital mortality} We follow a previously defined method for cohort selection in MIMIC~\cite{Harutyunyan2019}. We assume that the model has access to all notes charted during the first 48 hours of each patient's ICU stay. To avoid notes of poor semantic quality, we limit notes to the following types: 1) "Nursing"; 2) "Nursing/other"; and 3) "Physician". Since this task requires making predictions at the patient level, we concatenate note subsequences starting from the end of each patient's period of interest, working backwards, until we reach a limit of 30 subsequences, or exhaust all of a patient's notes. This is because notes written later during a patient's stay would be more informative than notes written at the beginning. This results in 15,892 total records with 55.3\% of them being male, 83.3\% English speakers, 80.9\% white, and 58.4\% using medicare.
    
\paragraph{Phenotyping using all notes} In addition to the cohort and note selection procedure above, we also add "Discharge summary" notes. The classification task is to predict patient membership in one of the 25 HCUP CCS code groups~\cite{Harutyunyan2019}, as linked by ICD-9 codes. We consider three tasks: 1) acute phenotype prediction; 2) chronic phenotype prediction; 3) all diseases. Therefore, this task actually consists of 28 separate binary classification problems of varying difficulties. Each task is composed of 30,598 samples, of which 56.07\% are male, 84.3\% are English speakers, 81.0\% are white, and 55.6\% using Medicare.
    
\paragraph{Phenotyping using first note} Following the same cohort selection, we select the first nursing or "Physician" note within the first 48 hours of a patient's stay. If this does not exist, we take the first "Nursing/other" note within the first 48 hours. If this also does not exist, the patient is dropped. This results in this set of tasks having different prevalences than the previous set. We use the same 28 binary classification tasks defined previously.

\subsection{Baseline Clinical BERT Pretraining}
\subsubsection*{Initialization}  Unlike previous approaches, we initialize our model from SciBERT, which has been shown to have better performance on a variety of benchmarking tasks~\cite{Beltagy2019}.

\subsubsection*{Tokenization and PHI Identifier Removal} Improving upon previous ClinicalBERT models, We make use of whole word masking, a recent development which resolves the issue of masked partial wordpiece tokens being too easy to predict~\cite{Cui2019}. We also replace the PHI identifiers with special tokens denoting the identifier type. For example, all date identifiers were replaced with a single special token for deidentified dates.  

\subsubsection*{Note Inclusion} We drop outpatient notes, as we cannot link these patients to demographic information such as insurance status. Notes were split into sentences and tokenized using the SciBERT tokenizer. Short sentences, which are common within the notes, were then aggregated with neighboring sentences into sequences of at least 20 tokens in length, to ensure that sequences fed into BERT during pretraining contain some level of semantic usefulness. 

\subsubsection*{Training} We first train one epoch (approximately 8 million samples) on sequences of combined length 128 using a batch size of 32, followed by one epoch (approximately 4 million samples) on sequences of combined length 512 using a batch size of 16. This model was trained on four GeForce GTX TITAN X 12 GB GPUs. 

\subsubsection*{Whole-note Embedding}
BERT has a fixed maximum input sequence length of 512 tokens. In order to fully capture the predictive power of notes that are longer than 512 tokens, we first split notes into subsequences of length 512, using a sliding window approach, up to a maximum of 10 subsequences. Then, we assign the outcome label for each subsequence to be the label from its derived note. We train our model to output probabilities at the subsequence level. Finally, for performance evaluations, we merge these probabilities to obtain the prediction for a note, using a previously proposed function~\cite{Huang2019}:

\begin{equation}
    P(Y=1) = \frac{P^n_{max} + P^n_{mean}n/c}{1+n/c}
\end{equation}

$P^n_{max}$ and  $P^n_{mean}$ are obtained by taking the maximum and mean of the probability outputs for the $n$ subsequences respectively. $c$ is a scaling factor which we tune for each task separately on the validation set.

\subsection{Note Template Generation}
We generate note templates based on the clinical notes in order to compare the likelihood of predicting a gender pronoun for the template in a fill-in-the-blanks task (described in Algorithm \ref{alg:logprob}). We consider topics in four categories for which biased treatments have been shown in health systems research: 1) chronic illnesses (heart disease, diabetes, hypertension); 2) mental health and addiction (addiction, mental illness, analgesics); 3) sexually-transmitted diseases (HIV); and 4) end-of-life treatment ("do not resuscitate" orders). Chronic illnesses often have more varied courses of treatment, which may be inadvertently affected by culture barriers and human bias~\cite{groce1993multiculturalism}. Mental health, addiction, and analgesic prescription have studied differences in access to treatment and other social determinants of health~\cite{sabshin1970dimensions}. HIV is a historically taboo condition in which homophobia and racism have studied effects on its treatment~\cite{arnold2014triply} "Do not resuscitate" orders have been show to have significant differences in its rate of assignment between African-Americans and white patients~\cite{shepardson1999racial}. 

\section{Methods}
In this work, we create our own baseline clinical BERT embedding model using all notes from the MIMIC \textit{NoteEvents} table. We train a classifier for each of the 57 finetuning tasks as described in Section \ref{sec:downstream_training}, and evaluate the classifier discrepancy gaps using the three definitions outlined in Section \ref{sec:quant_unfair}. We use bootstrapping of 1000 samples over the test set to establish a 95\% confidence interval for each gap, and report the total number of statistically significant gaps for each protected group.

\subsection{Downstream Training}
\label{sec:downstream_training}
For all fine-tuning tasks, we feed each subsequence into BERT in the sequence A position, and leave sequence B blank. We freeze the BERT weights and extract representations for each subsequence from BERT by concatenating the vectors of the last four hidden layers corresponding to the \texttt{[CLS]} tokens. This has been shown to give better performance than a variety of other contexual representation methods~\cite{devlin2019bert}, though it is still slightly worse than allowing backpropagation through the entire BERT model~\cite{devlin2019bert}. Since the goal of this work is not to obtain state-of-the-art performance in these tasks, but to instead to evaluate the bias in BERT representations, we did not want to alter these representations during fine-tuning.

To the extracted BERT representations, we also concatenate age, along with the OASIS, SAPS II and SOFA acuity scores~\cite{legallNewSimplifiedAcute1993, jonesSequentialOrganFailure2009, johnsonNewSeverityIllness2013}, which account for disease severity at admission.\footnote{The intuition is that very acute conditions often have more limited treatment options, which should be accounted for in a strong classifier.} We feed this vector into a fully connected neural network with batchnorm layers, ReLU activations, and ending in a sigmoid activation. A grid search is done over the number of layers, the ratio of the number of neurons in each layer to the previous, and the dropout rate. The model with the best AUPRC performance on the validation set is selected. For metrics that require a binary prediction (e.g. recall), we choose the threshold that results in the best F1 score on the validation set. We analyze bias for the following protected attributes: gender, language, ethnicity, insurance.

\subsubsection*{Train-Test Splits}
For all tasks, we use the same held-out test set defined in previous work~\cite{Harutyunyan2019}. We split 20\% of the remaining data as the validation set. No patient appears across the splits. In total, we have 57 downstream clinical tasks. For brevity in figures, we assign it a shorthand form, as shown in Table~\ref{tab:shorthand}, along with the overall prevalence for the \textit{phenotype using all notes} task. For the \textit{phenotyping using all notes} task, we prepend the shorthand with ``PA'', and for the \textit{phenotyping using the first note} task, we prepend the shorthand with ``PF''. See Appendix Table \ref{tab:tasks_prevelance} for a prevalence report of all tasks.

\begin{table*}[tbp!]
\begin{tabular}{|l|l|l|l|}
\hline
\textbf{Fairness Property}                                                         & \textbf{Definition}                                              & \textbf{Gap Name} & \textbf{Gap Equation}                                 \\ \hline
Demographic parity                                                                 & $P(\hat{Y}=y) = P(\hat{Y} = \hat{y} | Z=z),  \forall z \in Z $   & Parity Gap        & $\frac{TP_1 + FP_1}{N_1} - \frac{TP_2 + FP_2}{N_2}$   \\ \hline
\begin{tabular}[c]{@{}l@{}}Equality of opportunity\\ (positive class)\end{tabular} & $P(\hat{Y}=1 | Y=1) =  P(\hat{Y}=1 | Y=1, Z=z), \forall z \in Z$ & Recall Gap        & $\frac{TP_1 }{TP_1+FN_1} - \frac{TP_2 }{TP_2 + FN_2}$ \\ \hline
\begin{tabular}[c]{@{}l@{}}Equality of opportunity\\ (negative class)\end{tabular} & $P(\hat{Y}=0 | Y=0) =  P(\hat{Y}=0 | Y=0, Z=z), \forall z \in Z$ & Specificity Gap   & $\frac{TN_1 }{TN_1+FP_1} - \frac{TN_2 }{TN_2 + FP_2}$ \\
\hline
\end{tabular}
\caption{\label{tab:fairness_defs} The three fairness properties we will be using to evaluate downstream binary classifiers, their mathematical definitions, and the definition of the gap for the case with two protected groups. }
\end{table*}

\subsection{Evaluation of Classifier Fairness}
\label{sec:quant_unfair}
In this work, we evaluate classifiers using three oft-used definitions of fairness: 1) demographic parity~\cite{zemel2013learning}; 2) equality of opportunity for the positive class~\cite{hardt2016equality} and 3) or equality of opportunity for the negative class. To evaluate a fairness gap between two groups, we examine the difference between the relevant probabilities between said groups, as defined in Table \ref{tab:fairness_defs}.

In this work, we prioritize the recall gap, as it is the most clinically relevant definition of fairness. As machine learning models are most likely used as diagnostic tools, fewer false negatives is preferred to fewer false positives. Additionally, since all of our downstream clinical tasks have imbalance towards the negative class, an undesirable classifier that constantly predicts the majority (negative) class would still achieve 100\% true negative rate and have zero specificity gap.

We include demographic parity for comparison with other domains where it is a relevant metric of fairness. However, we note that it may be problematic in healthcare~\cite{hardt2016equality, Pfohl2019}, as optimizing for demographic parity might result in more bias elsewhere~\cite{Kleinberg2017}.

\subsubsection*{Multi-class Fairness Expansions}
\label{sec:multigroup}
For cases with more than two protected groups, we err on the side of caution and adopt the most aggressive definition of fairness for the healthcare setting. For a particular group, we report the maximum performance gap between said group and all other groups.

Consider a binary task $Y$, with protected attributes \\ $z=\{z_1, ... z_i, ... z_K\}$,\footnote{For example, the components of the protected attribute $z = $``gender" is $z_1 = $``male" and $z_2 = $``female".} each with number of people $n_i$, of which $c_i$ have been predicted to be condition positive by the classifier.

We follow methods used in prior work~\cite{hashimoto2018fairness} to expand the demographic parity gap, and use a similar process to obtain the recall and specificity gaps. We first define:
\[i^{\ast} = \argmax_{i\in z} \left|\frac{c_j}{n_j} - \frac{c_i}{n_i}\right|\]
Then the parity gap can be calculated by:
\begin{equation}
    gap_j = \frac{c_j}{n_j} - \frac{c_{i^{\ast}}}{n_{i^{\ast}}}
\end{equation}


\subsection{Evaluation of Log Probability Score}
We use \textit{log probability bias scores}, a previously proposed method for evaluating evaluating biases in contextual language models~\cite{kurita2019measuring}, which measures prior-adjusted likelihoods of predicting a given word for a fill-in-the-blanks task (Algorithm \ref{alg:logprob}).

For each topic, a set of template sentences are prepared, accounting for the various short-hands that different clinicians might adopt. For each template sentence, we compute log-probabilities for male words versus female pronouns. After running this for all sentences, we use a Wilcoxon signed-rank test~\cite{rosner2006wilcoxon}, to compare if the mean log probability bias scores have statistically significant differences for the male and female categories within a set of template sentences.

We expect that many conditions will not have equal base rates between men and women. Thus, a statistically significant log probability score does not necessarily imply unintended bias on the part of the model. To take this into account, we explore disease prevalences reported in the literature where possible. We also compute gender ratios within MIMIC-III. To do this, we first examine, out of all discharge notes, how many of them contain \textit{any} of the attribute strings for a particular topic. Then, we compute, out of all patients who do have the label, what percentage of them are male or female. 

\DontPrintSemicolon
\begin{algorithm}
\SetKwFunction{CalcLogScore}{CalcLogScore}
\SetKwProg{Fn}{Function}{:}{}
\SetKwData{Left}{$a_m$}
\SetKwData{This}{this}
\SetKwData{Up}{up}
\SetKwFunction{Extract}{GetProb}
\SetKwFunction{Insert}{Insert}
\SetKwFunction{Log}{log}
\SetKwFunction{Wilcoxon}{Wilcoxon}
 \KwData{$T$ - Set of template strings\\ $W_a$ - Set of strings for the attribute position
 \\ $W_{\mars}$ - Set of strings for men \\ $W_{\female}$ - Set of strings for women \\ $M$ - BERT model}
 
 \Fn{\CalcLogScore{$T$, $W_a$, $W_{\mars}$, $W_{\female}$, $M$}}{
 $y_{\mars},  y_{\female} \leftarrow \{\}, \{\}$ \;
 \For{$t\in T$}{
 $ind_a \leftarrow $ position in $t$ to insert attributes\;
 \For{$w_a \in W_a$ }{
 $t_{prior} \leftarrow $ \Insert $w_a$ into $t$ at $ind_a$ \;
 $ind_g \leftarrow $ position in $t_{prior}$ to insert the target \;
 \For{$i \in \{\mars, \female \}$}{
 \For{$w_{i} \in W_i$}{
  $p_{prior} \leftarrow$ \Extract{$M(t_{prior}), ind_g, w_{i}$}\;
  $t_{i} \leftarrow $ \Insert $w_{i}$ into $t_{prior}$ at $ind_g$ \;
  $p_{i} \leftarrow $ \Extract{$M(t_{i}), ind_g, w_{i}$} \;
  $y_{i} \cup $ \Log{$\frac{p_{i}}{p_{prior}}$} \;
 }
}
}
 }
 \KwRet{$y_{\mars}$, $y_{\female}$} \;
}
 \caption{Algorithm for computing prior-adjusted log probability bias scores, which we apply to a set of sentences for each medical context in Table \ref{tab:logprob}. We use the Wilcoxon Signed-Rank test to test for significant difference.}
\label{alg:logprob}
\end{algorithm}

\subsection{Algorithmic Debiasing}

\begin{figure}[]
    \centering
    \includegraphics[width=0.5\textwidth]{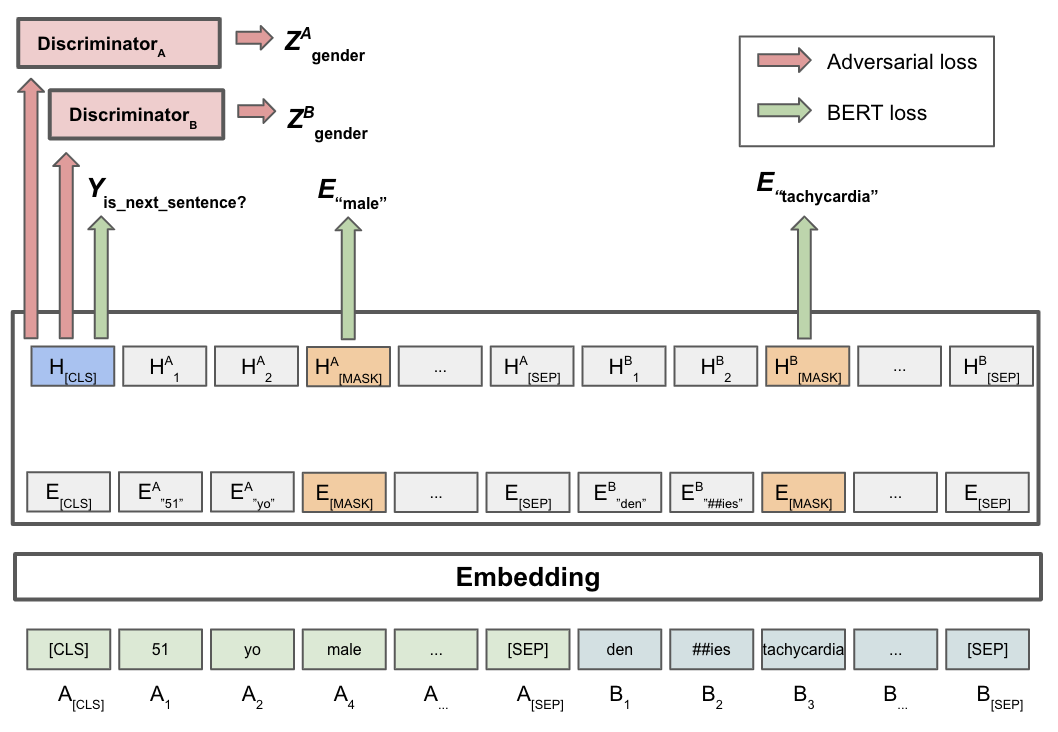}
    \caption{Procedure for adversarial debiasing. The gradients for the adversarial network (shown in red) are reversed during training, and combined with the loss functions form the BERT training proxy tasks.}
    \label{fig:pretrainDiagram}
\vspace{-1em}
\end{figure}

To explore how the baseline clinical BERT model can be debiased, we use an established adversarial debiasing approach~\cite{zhang2018mitigating, elazar2018adversarial, Beutel2017, Madras2018,edwards2015censoring} which uses a discriminator model to insist that the learned embedding be minimally predictive of protected subgroup information during pretraining, as described in Figure \ref{fig:pretrainDiagram}.
In this work, we examine only debiasing during pretraining, without access to any downstream labels. The motivation is that this would allow us to release a set of ``debiased'' BERT embeddings, similar to what has been done with non-contextual models~\cite{bolukbasi2016man}.  
We present our results for adversarial debiasing using gender as the protected group in Section \ref{subsection:debias}. Results for adversarial debiasing for language, ethnicity, and insurance are shown in Appendix \ref{subsection:other_groups}.
We use two (one for each BERT subsequence) three-layer fully connected, ReLU-activated neural network discriminator models. More details regarding our adversarial implementation can be found in Appendix \ref{sec:appendix_adversial}. 

\section{Results}
We find that our baseline clinical BERT model is biased via both the log probability bias score and performance discrepancy across groups in downstream tasks.

\subsection{Further Pretraining on Clinical Text Captures Gender Prevalence in Notes}

In Table \ref{tab:logprob}, we report log probability bias scores assessed on the publicly-available SciBERT model~\cite{Beltagy2019}, pretrained on biomedical text, and our baseline clinical BERT model, which further pretrains SciBERT on clinical notes. 

Pretraining on clinical notes integrates the inductive bias of the notes into the model. First, the model is more confident in its predictions. While only two out of eight note categories for SciBERT have a significant difference in gender, seven out of eight categories have a significant difference in baseline clinical BERT. This may be due to the fact that the note templates we use contain vocabulary that is much more likely appear in clinical notes than general scientific text. Second, further pretraining on clinical text shifts the model predictions towards the gender majority in the training data. Almost all of the conditions considered in Table \ref{tab:logprob} appear more frequently in the discharge summaries of males than females. After pretraining, model predictions shifts from predicting female-gendered pronouns slightly more frequently to almost exclusively predicting male-gendered pronouns. Pretraining on clinical notes effectively integrates gender-related associations from the notes into the model.


\begin{table*}[]
\resizebox{.99\textwidth}{!}{
\begin{tabular}{l|c|c|l|l|cll}
\cline{2-5}
\multicolumn{1}{c|}{\textbf{}}                  & \multicolumn{4}{c|}{\textbf{Log Probability Bias Scores}}                                                 & \multicolumn{1}{l}{}                                                                                      &                                                                                                              &                                                                                   \\ \cline{2-8} 
\multicolumn{1}{c|}{\multirow{2}{*}{\textbf{}}} & \multicolumn{2}{c|}{\textbf{SciBERT}} & \multicolumn{2}{l|}{\textbf{Clinical BERT}}                       & \multicolumn{1}{c|}{\multirow{2}{*}{\textbf{\begin{tabular}[c]{@{}c@{}}\# of \\ Templates\end{tabular}}}} & \multicolumn{1}{l|}{\multirow{2}{*}{\textbf{\begin{tabular}[c]{@{}l@{}}Gender Ratio\\ (M, F)\end{tabular}}}} & \multicolumn{1}{l|}{\multirow{2}{*}{\textbf{Sample Template}}}                    \\ \cline{2-5}
\multicolumn{1}{c|}{}                           & \textbf{M}        & \textbf{F}        & \multicolumn{1}{c|}{\textbf{M}} & \multicolumn{1}{c|}{\textbf{F}} & \multicolumn{1}{c|}{}                                                                                     & \multicolumn{1}{l|}{}                                                                                        & \multicolumn{1}{l|}{}                                                             \\ \hline
\multicolumn{1}{|l|}{Addiction}                 & 0.202             & 0.313             & 0.021*                          & -0.515*                         & \multicolumn{1}{c|}{2048}                                                                                 & \multicolumn{1}{l|}{57.4\%, 42.6\%}                                                                          & \multicolumn{1}{l|}{this is a 50 yo {[}GEND{]} with a hx of heroin addiction}     \\
\multicolumn{1}{|l|}{Heart Disease}             & 0.204*            & 0.333*            & 0.264*                          & -0.352*                         & \multicolumn{1}{c|}{18000}                                                                                & \multicolumn{1}{l|}{58.7\%, 41.3\%}                                                                          & \multicolumn{1}{l|}{this is a 82 yo {[}GEND{]} with a hx of cvd}                  \\
\multicolumn{1}{|l|}{Diabetes}                  & 0.100             & 0.251             & 0.205*                          & -0.865*                         & \multicolumn{1}{c|}{3600}                                                                                 & \multicolumn{1}{l|}{56.3\%, 43.7\%}                                                                          & \multicolumn{1}{l|}{this is a 45 yo {[}GEND{]} with a pmh of diabetes}            \\
\multicolumn{1}{|l|}{``Do Not Resuscitate''}    & 0.070             & 0.032             & -0.636*                         & -1.357*                         & \multicolumn{1}{c|}{256}                                                                                  & \multicolumn{1}{l|}{51.9\%, 48.1\%}                                                                          & \multicolumn{1}{l|}{{[}GEND{]} pt is dnr}                                         \\
\multicolumn{1}{|l|}{Analgesics}                & 1.295             & 2.127             & -0.077                          & 0.105                           & \multicolumn{1}{c|}{480}                                                                                  & \multicolumn{1}{l|}{56.9\%, 43.1\%}                                                                          & \multicolumn{1}{l|}{{[}GEND{]} is prescribed codeine}                             \\
\multicolumn{1}{|l|}{HIV}                       & 0.129             & 0.317             & 0.616*                          & -1.247*                         & \multicolumn{1}{c|}{3600}                                                                                 & \multicolumn{1}{l|}{64.6\%, 35.4\%}                                                                          & \multicolumn{1}{l|}{{[}GEND{]} has a pmh of hiv}                                  \\
\multicolumn{1}{|l|}{Hypertension}              & 0.413             & 0.437             & 0.440*                          & -0.402*                         & \multicolumn{1}{c|}{10800}                                                                                & \multicolumn{1}{l|}{55.8\%, 44.2\%}                                                                          & \multicolumn{1}{l|}{this is a 82 yo {[}GEND{]} with a discharge diagnosis of htn} \\
\multicolumn{1}{|l|}{Mental Illness}            & -0.414*           & -0.164*           & 0.084*                          & -0.263*                         & \multicolumn{1}{c|}{9000}                                                                                 & \multicolumn{1}{l|}{48.4\%, 51.6\%}                                                                          & \multicolumn{1}{l|}{this is a 45 yo {[}GEND{]} with a hx of schizophrenia}        \\ \hline
\end{tabular}
}
\caption{In the original SciBERT model, only 2/8 categories have a significantly different log probability score between genders. Baseline clinical BERT further trains SciBERT on medical notes, which shifts gender likelihood towards the majority group, creating a significant difference between the prior-adjusted likelihood of observing a gender for 7/8 medical context categories. ``Gender Ratio'' lists the gender composition of patients who have a \textit{positive} label, e.g., $57.4\%$ of all patients who have an ``Addiction'' label are men. \textit{*Denotes statistically significant difference between male and female at p < 0.01.}}
\label{tab:logprob}
\end{table*}

\subsection{Performance Gaps Favor the Majority Group}
We measure the performance gaps in our classifiers trained with baseline clinical BERT embeddings on 57 clinically relevant downstream tasks. 

\begin{figure*}[htbp!]
\centering
\begin{minipage}{0.47\textwidth}
\includegraphics[width=0.99\textwidth]{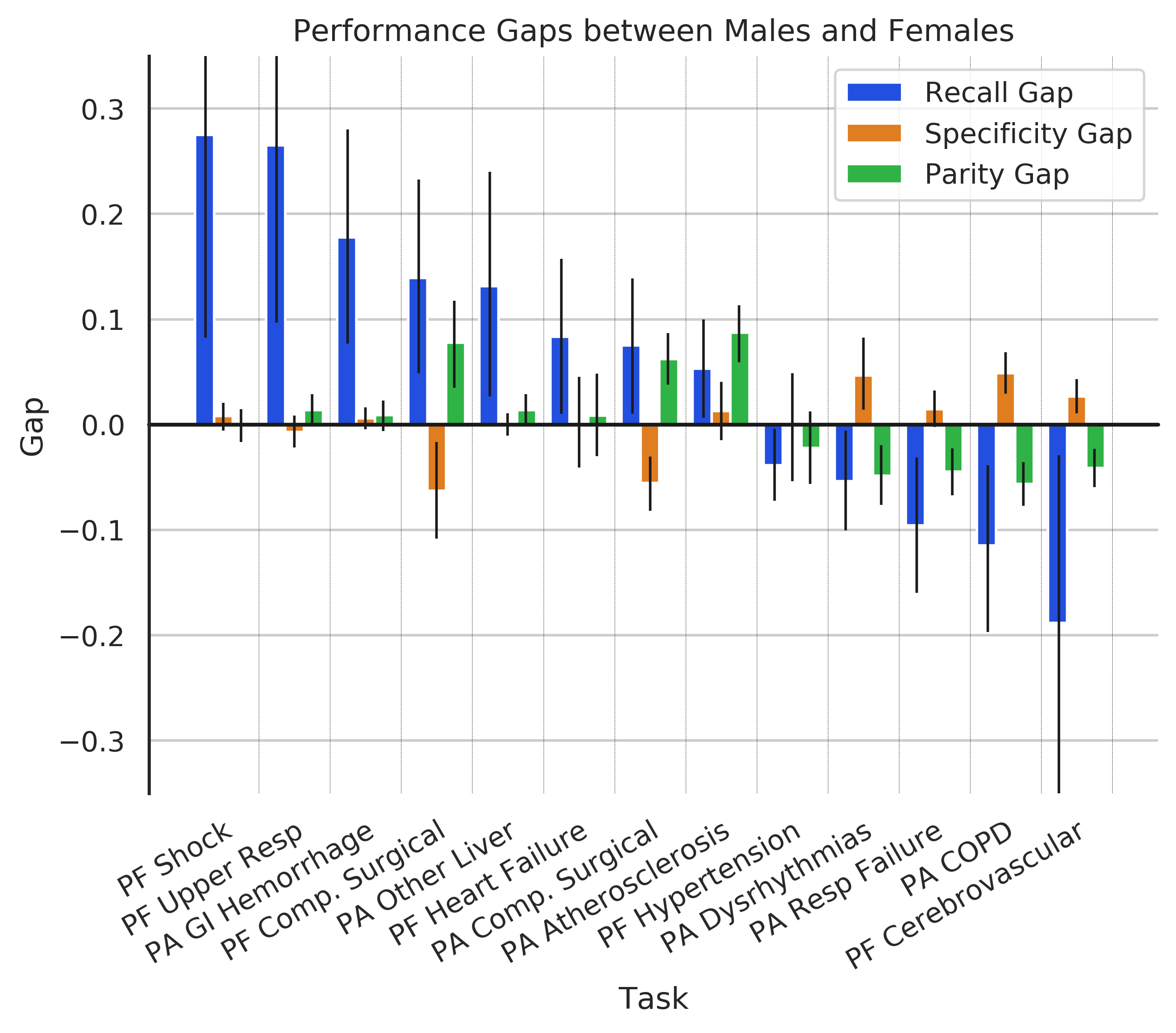}
\caption{Gendered performance gap measures for 13/57 clinical tasks with a significant recall gap between males and females. A positive bar indicates that the model performs better for \textit{females} than males; the recall gap generally favors men over women. }
\label{fig:gender_diagram}%

\end{minipage}
\hspace{0.05\textwidth}
\begin{minipage}{0.47\textwidth}
\includegraphics[width=0.99\textwidth]{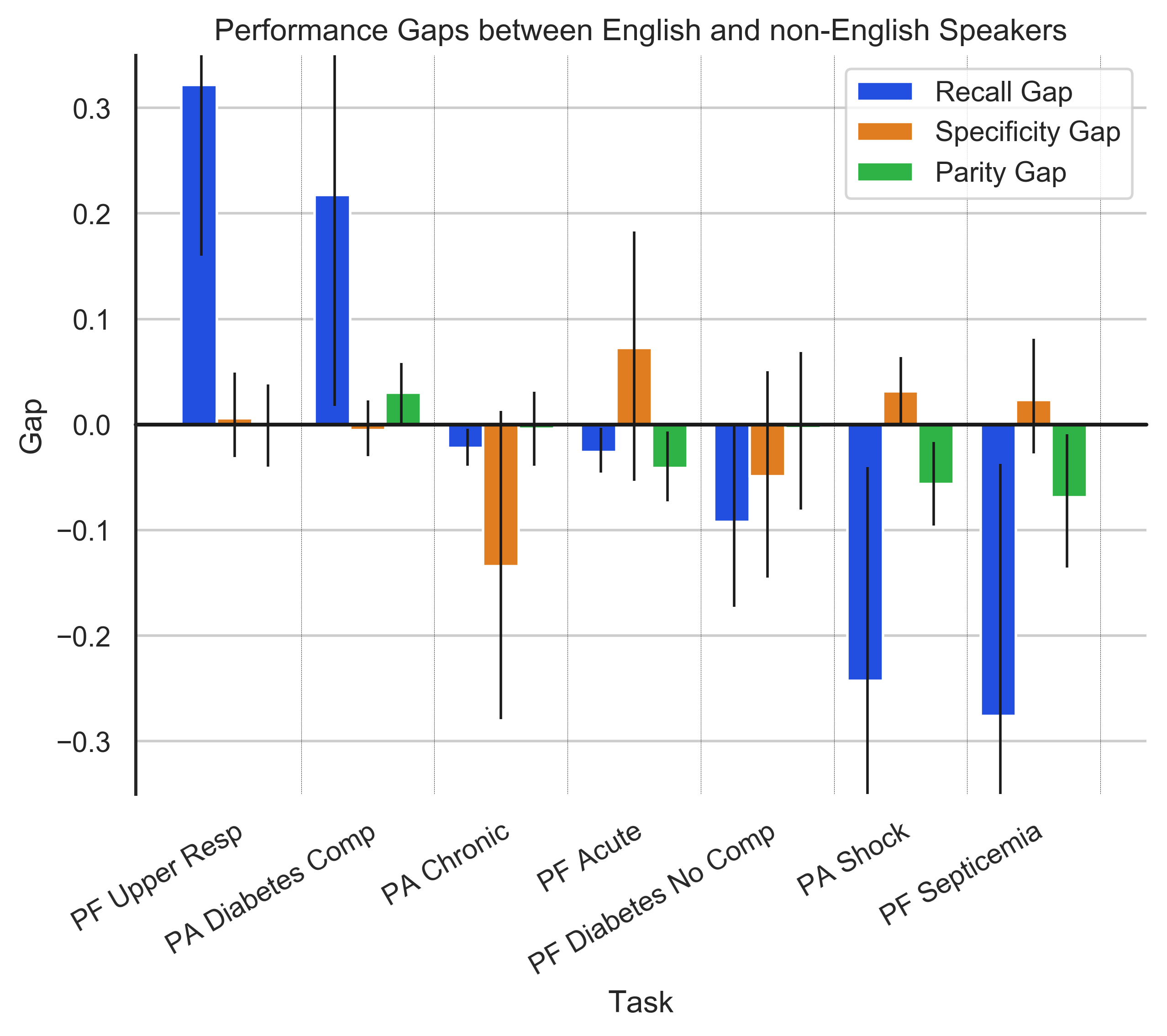}
\caption{Spoken language performance gap measures for 7/57 clinical tasks with a significant recall gap between English and non-English speakers. A positive bar indicates that the model performs better for \textit{English speakers} than non-English speakers; the recall gap generally favors non-English speakers over English speakers. }
\label{fig:language_diagram}%
\end{minipage}
\label{fig:ex3}%
\end{figure*}

First, we visualize the parity, recall, and specificity gap for gender (Figure \ref{fig:gender_diagram}) and language spoken (Figure \ref{fig:language_diagram}). Due to space limitations, we show tasks for which a significant difference exists for the recall gap, which we outline in Section \ref{sec:quant_unfair} as the most clinically-relevant definition. As shown, there are significant gaps in 13/57 and 7/57 of the downstream tasks respectively, at varying levels of disparity. 

We further compare the total number of significant differences in performance gaps for different genders, languages, ethnicities, and insurance types in Table \ref{tab:group_results}. Since performance gaps have directionality (i.e, which group performs better), we report these as parentheses for each subgroup. For the language and ethnicity attributes, individuals with unknown values were dropped when analyzing that group. The self-pay and government groups were dropped when analyzing insurance, due to their small relative class sample size. For multi-class comparisons (i.e., ethnicity and insurance), we present the results based on the multi-class fairness expansion as defined in Section~\ref{sec:quant_unfair}. 

For all protected class examinations in Table \ref{tab:group_results}, the recall gap favours (i.e. performs better for) the majority group, with the exception of language. Eight out of 13 significant tasks for gender favour "Male".
For ethnicity, the model often favor White and Asian patients when performance gaps exist, but seldomly favor Black and Hispanic patients. Further, we note that the model performs poorly for the "Other" ethnicity label when compared to other ethnicities in MIMIC-III. This category arises from poorly labeled data, and has high heterogeneity. For insurance subgroups, the model performance favours Medicare patients versus those under Medicaid and private insurance. 

We also perform multiple testing correction by using the Benjamini-Hochberg procedure to control for the false discovery rate (FDR)~\cite{benjamini1995controlling}, show in Appendix Table \ref{tab:group_results_hyp}. We find that many of the gaps, specifically relating to ethnicity and insurance, still remain.

\begin{table*}[htb!]
\begin{tabular}{ll|l|l|l|}
\cline{3-5}
                                                 &                                                    & \multicolumn{3}{l|}{\textbf{Significant Differences by Fairness Definition}} \\ \cline{3-5} 
                                                 &                                                    & \textbf{Recall Gap} & \textbf{Parity Gap }                 & \textbf{Specificity Gap}         \\ \hline
\multicolumn{1}{|l|}{\textbf{Gender}}                     & Male vs. Female (\% of Tasks Favoring Male)        & \textbf{13 (62\%)}           & 25 (36\%)        & 20 (80\%)               \\ \hline
\multicolumn{1}{|l|}{\textbf{Language}}                   & English vs. Other (\% of Tasks Favoring English)   & \textbf{7 (29\%)}           & 17 (12\%)        & 9 (89\%)               \\ \hline
\multicolumn{1}{|l|}{\multirow{5}{*}{\textbf{Ethnicity}}} & White vs. Other (\% of Tasks Favoring White)       & \textbf{4 (75\%)}         & 22 (82\%)        & 12 (17\%)                \\
\multicolumn{1}{|l|}{}                           & Black vs. Other (\% of Tasks Favoring Black)       & \textbf{5 (20\%)}           & 18 (72\%)            & 11 (18\%)               \\
\multicolumn{1}{|l|}{}                           & Hispanic vs. Other (\% of Tasks Favoring Hispanic) & \textbf{7 (0\%)}             & 18 (0\%)             & 20 (100\%)              \\
\multicolumn{1}{|l|}{}                           & Asian vs. Other (\% of Tasks Favoring Asian)       & \textbf{8 (62\%)}           & 7 (100\%)           & 8 (50\%)               \\
\multicolumn{1}{|l|}{}                           & "Other" vs. Other (\% of Tasks Favoring "Other")   & \textbf{10 (0\%)}             & 8 (0\%)           & 9 (100\%)              \\ \hline
\multicolumn{1}{|l|}{\multirow{3}{*}{\textbf{Insurance}}} & Medicare vs. Other (\% of Tasks Favoring Medicare) & \textbf{33 (85\%)}           & 51 (92\%)           & 48 (6\%)                \\
\multicolumn{1}{|l|}{}                           & Private vs. Other (\% of Tasks Favoring Private)   & \textbf{15 (7\%)}            & 41 (2\%)           & 40 (98\%)               \\
\multicolumn{1}{|l|}{}                           & Medicaid vs. Other (\% of Tasks Favoring Medicaid) & \textbf{20 (20\%)}           & 31 (19\%)           & 30 (83\%)               \\ \hline

\end{tabular}
\caption{Classifiers trained with baseline clinical BERT embeddings have multi-group fairness performance gaps (defined in Section \ref{sec:multigroup}) across gender, language, ethnicity, and insurance status. We count number of downstream classification tasks with a statistically significant differences (out of 57 total), as well as the percentage of significant tasks which \textit{favor} a subgroup. For many comparisons, there are a large number of tasks for which a significant difference exists across subgroups. We list the recall gap first, as it is the most aggressive quantification for this high-stake application (Section \ref{sec:quant_unfair}).}
\label{tab:group_results} 
\end{table*}

\subsection{Adversarial Debiasing During Pretraining Is Insufficient}
\label{subsection:debias}
In Table \ref{tab:origvsdebiasgender_bias}, we present a comparison of the classifiers' performance from embeddings generated using the baseline clinical BERT against that of the debiased clinical BERT. We find that the debiased clinical BERT produce strong downstream performance overall, retaining a correlation of 0.996 in AUPRC with the baseline clinical BERT (Appendix Figure \ref{fig:auprcs}, Appendix Table \ref{tab:tasks_performance}). Results in the Appendix Table \ref{tab:predict_groups} further outline the performance of the adversarial model and its effects on disparity. 

While debiasing slightly reduces the number of tasks for which there is a significant recall gap, this reduction is insufficient for deployment in a high-stake medical setting. The debiasing also does not seem to reduce the specificity gap. Results for adversarial debiasing during pretraining with language, ethnicity, and insurance as protected groups show similar results, and are presented in Appendix Section \ref{tab:group_results}. Additionally, although the adversary is able to fool the discriminator during adversarial training, a classifier applied post-hoc can still extract information about sensitive attributes (Appendix Table \ref{tab:predict_groups}), reaffirming past work \cite{elazar2018adversarial}.

\begin{table}[h!]
\begin{tabular}{|l|l|l|l|}
\hline
\multicolumn{4}{|c|}{\textbf{Significant Gap Count (\% Favoring Male)}}\\ \hline
\textbf{Model} & \textbf{Parity Gap} & \textbf{Recall Gap} & \textbf{Specificity Gap} \\ \hline
Baseline & 25 (36\%)           & \textbf{13 (62\%)}  & 20 (80\%)                \\ \hline
Debiased & 25 (36\%)           & \textbf{9 (56\%)}  & 20 (70\%) \\
\hline
\end{tabular}
\caption{Comparison of classifiers based on our original clinical BERT and the gender-debiased clinical BERT on 57 tasks. Significant performance gaps across males versus females are shown. Note that the ``debiasing'' does not greatly reduce the number of statistically significant gaps.
\label{tab:origvsdebiasgender_bias}}
\end{table}

\section{Discussion}

\subsection{Further Pretraining Captures Note Biases}
As seen in Table \ref{tab:logprob}, further pretraining SciBERT on clinical notes creates a model that is significantly more likely to predict male gendered pronouns. This matches the gender ratio of the conditions we probe in the MIMIC notes, but integrating the inductive bias of the notes into the model could also bring unintended biases. Hypertension is an example of an examined category where the real-world disease prevalence is roughly equal for men and women~\cite{yoon2015hypertension}. However, hypertension-related words appear more frequently in the discharge notes of males than females, and our baseline clinical BERT is significantly more likely to predict patient gender to be male after pretraining on these notes. This side-effect of learning unwanted latent relationships from data should be carefully audited -- and in some cases, debiased -- to avoid propagating unwanted effects by machine learning systems.

\subsection{Fairness Gaps Along Known Biases}
We note that some of the biases which we quantify may be reflective of known biases present in our medical system. For example, males tend to have higher recall than females for most tasks (eight out of 13 tasks), which could be due to women experiencing more complex co-morbidities, thus rendering it more difficult to correctly predict any single disease~\cite{hwang2001out}. Furthermore, we note that the higher prevalence of heart disease for males in MIMIC-III (Table \ref{tab:logprob}) accords with previous studies on the under-diagnosis of heart disease in women~\cite{milner2004gender}. 
Similarly, we see that patients of Black and Hispanic/Latino descent suffer from lower recall than other groups. This observed bias may be due to under-utilization of the healthcare system~\cite{dubard2006effect, trent2019barriers, fiscella2002disparities}. 

Overall, performance gaps of the baseline clinical BERT model on downstream tasks illustrate the need for algorithmic auditing and debiasing before the deployment of such models. For the ``Medicare vs. Other" performance comparison, over half of all tasks had a significant difference. This demonstrates that performance disparities across demographic subgroups are a relevant consideration for a variety of real-world applications for such models.
While we do not believe that each of the performance gaps will directly relate to treatment and outcome disparities observed in the real world, it is important to note that disentangling performance differences due to true medical confounders and unwanted biases is extremely difficult, if not impossible. Our results demonstrate the need for further research into methods to identify and remove disparities before any model is considered for deployment.

\subsection{Challenges in Using Adversarial Debiasing for Clinical Contextual Embedding Models}
As seen in Table \ref{tab:origvsdebiasgender_bias}, debiasing during pretraining does not greatly reduce fairness gaps compared to the the baseline clinical BERT model. For a high capacity encoder like BERT, the adversarial decoder might be ``underpowered". However, it is unfeasible to train a discriminator model of similar capacity to the overall BERT model, which raises a central issue in using adversarial methods to debias contextual embeddings. It is also important to consider downstream uses of BERT embeddings; applying adversarial debiasing to only the \texttt{[CLS]} token would in theory debias classification tasks, but not sequence-based tasks such as named entity recognition (NER) or question answering. Finally, adversarial debiasing during pretraining might not be conceptually desirable in the first place. If the model is encouraged to not encode information about the protected group $Z$ in the representation, the predicted label $\hat{Y}$ would be independent of $Z$, resulting in demographic parity~\cite{zhang2018mitigating, Beutel2017}, which, as previously discussed, is a problematic definition in healthcare. Adversarial debiasing during finetuning would theoretically allow other definitions of fairness to be achieved~\cite{zhang2018mitigating}, and is a possible direction for future works.

\section{Limitations and Future Work}

There are limitations to our work that provide opportunity for future efforts. First, our downstream tasks are derived from ICD-9 labels, which are assigned post-stay by the hospital billing department. We treat these labels as the gold standard, and disregard the possibility that there may be errors or biases in the labels themselves. Second, we do not address intersectional discrimination (e.g. gender and race), which has been shown to important for auditing algorithmic fairness~\cite{buolamwini2018gender}. Third, we make no attempt to disentangle the sources of bias in our models, whether it be data imbalance, data quality issues, or inherent social biases in the healthcare system. Fourth, we do not attempt to debias the data itself (whether it be pretraining or finetuning) directly. As shown here and in other work~\cite{elazar2018adversarial, gonen2019lipstick}, debiasing methods might not always be effective and might simply hide the bias instead of removing it~\cite{gonen2019lipstick}. Automated data augmentation methods might work in some cases~\cite{iosifidisdealing}, but building a high-quality unbiased set of training data (for example, by removing biased documents) can be time consuming~\cite{brunetUnderstandingOriginsBias2018}. Finally, it is unclear whether it would be ethical to use a ``fairer'' model which trades off model performance (and thereby patient outcomes) in one group for another in a healthcare setting. These are decisions which would have to be made on a case-by-case basis by healthcare providers.


\section{Conclusion}

When a machine learning algorithm is trained on data that is fundamentally biased, it can result in a model that reflects (or even amplifies) those biases~\cite{mehrabi2019survey}. In a high-stakes setting like healthcare, model biases must be even more carefully examined. There are several ways in which data can become biased, ranging from intentional societal discrimination or stereotyping, to simple group imbalances~\cite{rajkomar2018ensuring}. Before deploying a model, especially in healthcare, its biases should be carefully examined across protected groups and intersections of protected groups~\cite{mitchell2019model}. 

In this work, we pretrain BERT on a large corpora of medical notes, and using quantitative and qualitative methods, demonstrate that these embeddings propagate unwanted latent relationships with regards to different genders, language speakers, ethnicities, and insurance groups. Using different fairness definitions, we quantitatively demonstrate that significant differences exist in model performance for different groups. We also qualitatively examine the course of action which a BERT model trained in scientific text produces, when the medical context is kept the same, and the race is changed. Finally, we calculate log probability bias scores of filling in the gender pronoun of a note for males and females, and find that after further pretraining medical notes, our baseline clinical BERT model becomes more confident in the gender of the note, and may have captured relationships between gender and medical conditions which exceed biological associations.

We believe that our demonstration of this risk further cements the need for specialized algorithms that can detect and minimize the impact of such biases. With the growing interest in BERT and contextualized word embeddings, and the potentials of using machine learning in clinical settings, we encourage the machine learning community to explore fair contextual word embedding methods specific to the needs of the healthcare domain, and move towards fair, life-saving machine learning systems.






\begin{acks}
The authors would like to thank Alan Moses, Alex Lu, David Madras, Emily Alsentzer, Nathan Ng,  Quaid Morris, and Tristan Naumann for their helpful suggestions. Amy X. Lu is funded by the NSERC Canada Graduate Scholarship Master's award. Mohamed Abdalla is funded by the Vanier Canada Graduate Scholarship. Dr. Marzyeh Ghassemi is funded in part by Microsoft Research, a CIFAR AI Chair at the Vector Institute, a Canada Research Council Chair, and an NSERC Discovery Grant. We thank the Vector Institute for providing the computing resources for this study.
\end{acks}

\clearpage
\bibliographystyle{ACM-Reference-Format}
\bibliography{references}

\newpage
\appendix

\newpage \newpage \clearpage
\onecolumn
\setcounter{table}{0}
\renewcommand{\thetable}{\Alph{section}\arabic{table}}
\setcounter{figure}{0}
\renewcommand{\thefigure}{\Alph{section}\arabic{figure}}
\section{Appendix A - Descriptive Statistics and Model Performance}
\begin{table*}[h!]
\resizebox{\textwidth}{!}{%
\begin{tabular}{l|l|l|l|l|l|l}
\multicolumn{2}{c|}{\textbf{Task}}                                                                                 & \textbf{\# Samples}     & \textbf{\% Male}          & \textbf{Overall Prevalence} & \textbf{Prevalence in Males} & \textbf{Prevalence in Females} \\ \hline
                                  & In-hospital Mortality                                                           & 15892                   & 55.32\%                   & 13.21\%                     & 13.62\%                      & 12.89\%                        \\ \hline
\multirow{28}{*}{Phenotype All}   & Acute and unspecified renal failure                                            & \multirow{28}{*}{30598} & \multirow{28}{*}{56.07\%} & 20.10\%                     & 19.93\%                      & 20.23\%                        \\
                                  & Acute cerebrovascular disease                                                  &                         &                           & 7.08\%                      & 7.98\%                       & 6.38\%                         \\
                                  & Acute myocardial infarction                                                    &                         &                           & 11.43\%                     & 10.50\%                      & 12.15\%                        \\
                                  & Cardiac dysrhythmias                                                           &                         &                           & 31.56\%                     & 30.76\%                      & 32.18\%                        \\
                                  & Chronic kidney disease                                                         &                         &                           & 11.42\%                     & 10.21\%                      & 12.36\%                        \\
                                  & Chronic obstructive pulmonary disease and bronchiectasis                       &                         &                           & 12.91\%                     & 13.89\%                      & 12.15\%                        \\
                                  & Complications of surgical procedures or medical care                           &                         &                           & 20.40\%                     & 19.90\%                      & 20.79\%                        \\
                                  & Conduction disorders                                                           &                         &                           & 6.76\%                      & 6.08\%                       & 7.29\%                         \\
                                  & Congestive heart failure; nonhypertensive                                      &                         &                           & 27.73\%                     & 30.24\%                      & 25.76\%                        \\
                                  & Coronary atherosclerosis and other heart disease                               &                         &                           & 32.66\%                     & 25.80\%                      & 38.03\%                        \\
                                  & Diabetes mellitus with complications                                           &                         &                           & 9.09\%                      & 9.40\%                       & 8.84\%                         \\
                                  & Diabetes mellitus without complication                                         &                         &                           & 18.88\%                     & 18.74\%                      & 19.00\%                        \\
                                  & Disorders of lipid metabolism                                                  &                         &                           & 25.87\%                     & 23.72\%                      & 27.56\%                        \\
                                  & Essential hypertension                                                         &                         &                           & 41.07\%                     & 42.38\%                      & 40.04\%                        \\
                                  & Fluid and electrolyte disorders                                                &                         &                           & 23.67\%                     & 26.50\%                      & 21.45\%                        \\
                                  & Gastrointestinal hemorrhage                                                    &                         &                           & 7.31\%                      & 6.81\%                       & 7.71\%                         \\
                                  & Hypertension with complications and secondary hypertension                     &                         &                           & 11.99\%                     & 11.26\%                      & 12.56\%                        \\
                                  & Other liver diseases                                                           &                         &                           & 7.76\%                      & 6.84\%                       & 8.48\%                         \\
                                  & Other lower respiratory disease                                                &                         &                           & 4.11\%                      & 4.73\%                       & 3.63\%                         \\
                                  & Other upper respiratory disease                                                &                         &                           & 3.71\%                      & 4.23\%                       & 3.31\%                         \\
                                  & Pleurisy; pneumothorax; pulmonary collapse                                     &                         &                           & 8.50\%                      & 8.73\%                       & 8.32\%                         \\
                                  & Pneumonia (except that caused by tuberculosis or sexually transmitted disease) &                         &                           & 13.96\%                     & 14.49\%                      & 13.54\%                        \\
                                  & Respiratory failure; insufficiency; arrest (adult)                             &                         &                           & 17.53\%                     & 19.23\%                      & 16.20\%                        \\
                                  & Septicemia (except in labor)                                                   &                         &                           & 14.03\%                     & 14.49\%                      & 13.66\%                        \\
                                  & Shock                                                                          &                         &                           & 7.21\%                      & 7.63\%                       & 6.89\%                         \\
                                  & Any Chronic                                                                    &                         &                           & 78.63\%                     & 78.31\%                      & 78.87\%                        \\
                                  & Any Acute                                                                      &                         &                           & 78.56\%                     & 80.62\%                      & 76.95\%                        \\
                                  & Any Disease                                                                    &                         &                           & 92.68\%                     & 93.16\%                      & 92.29\%                        \\ \hline
\multirow{28}{*}{Phenotype First} & Acute and unspecified renal failure                                            & \multirow{28}{*}{16689} & \multirow{28}{*}{56.91\%} & 14.72\%                     & 15.39\%                      & 14.20\%                        \\
                                  & Acute cerebrovascular disease                                                  &                         &                           & 5.87\%                      & 6.73\%                       & 5.22\%                         \\
                                  & Acute myocardial infarction                                                    &                         &                           & 10.21\%                     & 8.82\%                       & 11.27\%                        \\
                                  & Cardiac dysrhythmias                                                           &                         &                           & 27.32\%                     & 26.16\%                      & 28.21\%                        \\
                                  & Chronic kidney disease                                                         &                         &                           & 10.80\%                     & 10.11\%                      & 11.33\%                        \\
                                  & Chronic obstructive pulmonary disease and bronchiectasis                       &                         &                           & 11.22\%                     & 12.02\%                      & 10.61\%                        \\
                                  & Complications of surgical procedures or medical care                           &                         &                           & 16.39\%                     & 15.76\%                      & 16.87\%                        \\
                                  & Conduction disorders                                                           &                         &                           & 6.27\%                      & 5.80\%                       & 6.63\%                         \\
                                  & Congestive heart failure; nonhypertensive                                      &                         &                           & 22.27\%                     & 24.74\%                      & 20.40\%                        \\
                                  & Coronary atherosclerosis and other heart disease                               &                         &                           & 33.42\%                     & 25.41\%                      & 39.48\%                        \\
                                  & Diabetes mellitus with complications                                           &                         &                           & 8.78\%                      & 9.29\%                       & 8.40\%                         \\
                                  & Diabetes mellitus without complication                                         &                         &                           & 18.37\%                     & 18.22\%                      & 18.48\%                        \\
                                  & Disorders of lipid metabolism                                                  &                         &                           & 28.78\%                     & 25.71\%                      & 31.10\%                        \\
                                  & Essential hypertension                                                         &                         &                           & 42.21\%                     & 42.73\%                      & 41.81\%                        \\
                                  & Fluid and electrolyte disorders                                                &                         &                           & 19.46\%                     & 22.18\%                      & 17.40\%                        \\
                                  & Gastrointestinal hemorrhage                                                    &                         &                           & 6.78\%                      & 6.51\%                       & 6.98\%                         \\
                                  & Hypertension with complications and secondary hypertension                     &                         &                           & 11.44\%                     & 11.15\%                      & 11.67\%                        \\
                                  & Other liver diseases                                                           &                         &                           & 6.22\%                      & 5.33\%                       & 6.90\%                         \\
                                  & Other lower respiratory disease                                                &                         &                           & 3.59\%                      & 4.28\%                       & 3.06\%                         \\
                                  & Other upper respiratory disease                                                &                         &                           & 2.54\%                      & 3.05\%                       & 2.16\%                         \\
                                  & Pleurisy; pneumothorax; pulmonary collapse                                     &                         &                           & 5.79\%                      & 5.87\%                       & 5.73\%                         \\
                                  & Pneumonia (except that caused by tuberculosis or sexually transmitted disease) &                         &                           & 7.60\%                      & 8.22\%                       & 7.13\%                         \\
                                  & Respiratory failure; insufficiency; arrest (adult)                             &                         &                           & 8.33\%                      & 9.32\%                       & 7.59\%                         \\
                                  & Septicemia (except in labor)                                                   &                         &                           & 8.81\%                      & 9.41\%                       & 8.36\%                         \\
                                  & Shock                                                                          &                         &                           & 3.81\%                      & 4.51\%                       & 3.28\%                         \\
                                  & Any Chronic                                                                    &                         &                           & 77.36\%                     & 76.26\%                      & 78.20\%                        \\
                                  & Any Acute                                                                      &                         &                           & 70.68\%                     & 72.84\%                      & 69.04\%                        \\
                                  & Any Disease                                                                    &                         &                           & 89.72\%                     & 90.03\%                      & 89.49\%                       
\end{tabular}
}
\caption{\label{tab:tasks_prevelance} Summary table of the number of samples, \% majority group for gender, and by-gender prevalence for all Healthcare Cost and Utilization Project clinical classifications software (HCUP CCS) disease groupings.}
\end{table*}

\begin{table*}[h!]
\resizebox{\textwidth}{!}{%
\begin{tabular}{l|l|l|l|l|l}
\multicolumn{2}{c|}{\textbf{Task}}                                                                                    & \textbf{Baseline AUROC} & \textbf{Baseline AUPRC} & \textbf{Debiased AUROC} & \textbf{Debiased AUPRC} \\ \hline
                                   & In-hospital Mortality &         86.74\% &         50.17\% &         85.96\% &         48.55\%                 \\ \hline
\multirow{28}{*}{Phenotype All}   & Acute and unspecified renal failure &         83.64\% &         57.89\% &         83.37\% &         57.87\% \\
                & Acute cerebrovascular disease &         93.51\% &         61.97\% &         92.60\% &         61.87\% \\
                & Acute myocardial infarction &         88.16\% &         63.29\% &         88.27\% &         63.42\% \\
                & Cardiac dysrhythmias &         77.87\% &         64.05\% &         77.02\% &         63.28\% \\
                & Chronic kidney disease &         86.06\% &         53.83\% &         86.08\% &         54.17\% \\
                & Chronic obstructive pulmonary disease and bronchiectasis &         75.36\% &         34.60\% &         76.39\% &         37.77\% \\
                & Complications of surgical procedures or medical care &         73.80\% &         41.87\% &         74.37\% &         41.59\% \\
                & Conduction disorders &         82.77\% &         38.81\% &         80.27\% &         26.86\% \\
                & Congestive heart failure; nonhypertensive &         83.08\% &         67.04\% &         83.16\% &         66.52\% \\
                & Coronary atherosclerosis and other heart disease &         86.67\% &         80.70\% &         86.45\% &         79.73\% \\
                & Diabetes mellitus with complications &         80.55\% &         45.30\% &         81.72\% &         47.76\% \\
                & Diabetes mellitus without complication &         66.65\% &         29.34\% &         67.04\% &         31.66\% \\
                & Disorders of lipid metabolism &         76.63\% &         51.48\% &         76.35\% &         51.06\% \\
                & Essential hypertension &         71.11\% &         59.47\% &         70.99\% &         59.87\% \\
                & Fluid and electrolyte disorders &         74.51\% &         46.22\% &         74.30\% &         46.56\% \\
                & Gastrointestinal hemorrhage &         87.28\% &         50.36\% &         88.14\% &         49.95\% \\
                & Hypertension with complications and secondary hypertension &         84.53\% &         51.67\% &         84.34\% &         50.18\% \\
                & Other liver diseases &         86.93\% &         53.41\% &         87.18\% &         53.07\% \\
                & Other lower respiratory disease &         65.45\% &          8.65\% &         64.26\% &          9.07\% \\
                & Other upper respiratory disease &         84.75\% &         42.13\% &         84.21\% &         43.74\% \\
                & Pleurisy; pneumothorax; pulmonary collapse &         69.63\% &         21.85\% &         71.02\% &         22.26\% \\
                & Pneumonia (except that caused by tuberculosis or sexually transmitted disease) &         81.66\% &         43.43\% &         82.13\% &         42.74\% \\
                & Respiratory failure; insufficiency; arrest (adult) &         89.98\% &         67.99\% &         90.01\% &         68.72\% \\
                & Septicemia (except in labor) &         87.85\% &         57.41\% &         87.14\% &         56.68\% \\
                & Shock &         87.66\% &         44.26\% &         88.44\% &         44.23\% \\
                & any acute &         82.29\% &         94.26\% &         82.14\% &         94.13\% \\
                & any chronic &         87.55\% &         95.87\% &         87.42\% &         95.80\% \\
                & any disease &         91.63\% &         99.25\% &         91.59\% &         99.24\% \\ \hline
\multirow{28}{*}{Phenotype First} & Acute and unspecified renal failure &         77.85\% &         38.65\% &         78.78\% &         40.24\% \\
                & Acute cerebrovascular disease &         83.81\% &         32.46\% &         86.41\% &         34.72\% \\
                & Acute myocardial infarction &         84.43\% &         52.38\% &         84.08\% &         50.53\% \\
                & Cardiac dysrhythmias &         70.43\% &         46.68\% &         69.86\% &         45.17\% \\
                & Chronic kidney disease &         82.31\% &         46.55\% &         81.54\% &         41.65\% \\
                & Chronic obstructive pulmonary disease and bronchiectasis &         69.46\% &         21.55\% &         70.57\% &         21.70\% \\
                & Complications of surgical procedures or medical care &         66.16\% &         27.21\% &         64.23\% &         24.46\% \\
                & Conduction disorders &         73.28\% &         18.75\% &         74.56\% &         20.22\% \\
                & Congestive heart failure; nonhypertensive &         77.10\% &         50.74\% &         76.20\% &         48.25\% \\
                & Coronary atherosclerosis and other heart disease &         82.75\% &         71.90\% &         81.77\% &         71.30\% \\
                & Diabetes mellitus with complications &         70.25\% &         26.85\% &         72.37\% &         29.77\% \\
                & Diabetes mellitus without complication &         63.61\% &         24.71\% &         64.47\% &         26.24\% \\
                & Disorders of lipid metabolism &         73.90\% &         51.31\% &         72.64\% &         49.90\% \\
                & Essential hypertension &         67.49\% &         56.83\% &         66.93\% &         56.38\% \\
                & Fluid and electrolyte disorders &         71.47\% &         38.50\% &         72.53\% &         38.62\% \\
                & Gastrointestinal hemorrhage &         77.66\% &         35.62\% &         76.89\% &         32.87\% \\
                & Hypertension with complications and secondary hypertension &         77.81\% &         40.72\% &         77.56\% &         38.14\% \\
                & Other liver diseases &         78.18\% &         33.04\% &         77.29\% &         33.59\% \\
                & Other lower respiratory disease &         55.68\% &          5.04\% &         57.83\% &          4.89\% \\
                & Other upper respiratory disease &         66.83\% &         11.75\% &         59.74\% &          8.46\% \\
                & Pleurisy; pneumothorax; pulmonary collapse &         55.86\% &         10.29\% &         55.57\% &          9.62\% \\
                & Pneumonia (except that caused by tuberculosis or sexually transmitted disease) &         70.65\% &         20.74\% &         71.66\% &         20.30\% \\
                & Respiratory failure; insufficiency; arrest (adult) &         81.83\% &         32.51\% &         81.53\% &         34.00\% \\
                & Septicemia (except in labor) &         82.62\% &         36.97\% &         83.66\% &         40.41\% \\
                & Shock &         84.05\% &         30.14\% &         84.14\% &         30.97\% \\
                & any acute &         74.75\% &         87.28\% &         73.67\% &         86.47\% \\
                & any chronic &         84.38\% &         94.30\% &         84.46\% &         93.90\% \\
                & any disease &         87.54\% &         98.24\% &         87.29\% &         98.23\%             
\end{tabular}
}
\caption{\label{tab:tasks_performance} Summary table of the predictive performance (evaluated by AUROC and AUPRC) of our trained clinical BERT model and the debiased model (with gender as the protected group). From embeddings generated from the respective BERT model, the task is to correctly classify the disease for all diseases in HCUP CCS groupings.}
\end{table*}

\begin{figure*}[h!]
    \centering
    \includegraphics[width=0.75\textwidth]{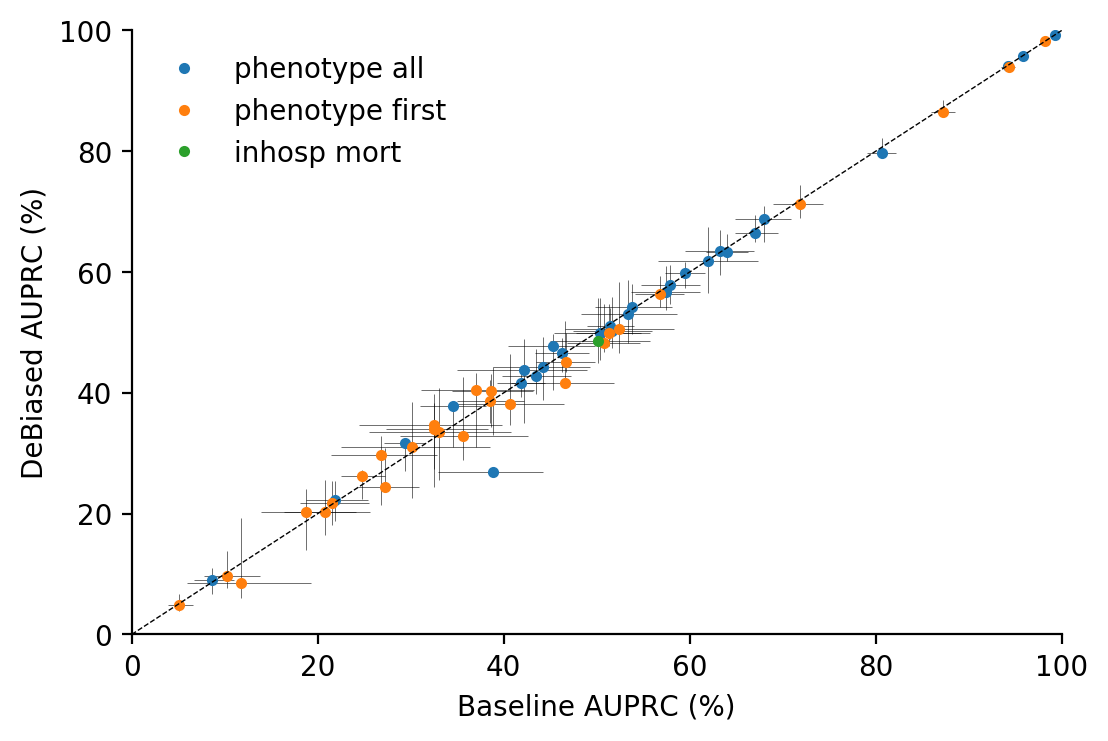}
    \caption{AUPRC for the baseline model evaluated on all downstream clinical tasks versus AUPRC for the debiased model. Error bars are 95\% confidence intervals from bootstrapping.}
    \label{fig:auprcs}
\end{figure*}

\clearpage

\section{Procedure for generating sentence completion figure}
\label{sec:fill_in_blank_procedure}
We create template sentences adapted from real medical notes in MIMIC-III, and asks SciBERT to predict the next two tokens, which corresponds to course of action. 
We then modify replace the ``[**RACE**]" token with races from the list [``caucasian", ``white", ``african", ``african american", ``black"], and see if SciBERT generates different predictions for the masked tokens. In the figure, we denote the word or word-pieces for BERT to predict with ``[**TOKEN**]". By convention, when generating predictions, we also add the ``[CLS]" and ``[SEP]" tokens to the beginning and end of the sentence, respectively, though we omit these tokens for the sake of clarity in figure. \ref{fig:fillinblanks}. 

\section{Multiple Hypothesis Correction of Table \ref{tab:group_results}}
\begin{table*}[htb!]
\begin{tabular}{ll|l|l|l|}
\cline{3-5}
                                                 &                                                    & \multicolumn{3}{l|}{\textbf{Significant Differences by Fairness Definition}} \\ \cline{3-5} 
                                                 &                                                    & \textbf{Recall Gap} & \textbf{Parity Gap }                 & \textbf{Specificity Gap}         \\ \hline
\multicolumn{1}{|l|}{\textbf{Gender}}                     & Male vs. Female (\% of Tasks Favoring Male)        & \textbf{6 (30\%)}           & 23 (67\%)        & 16 (81\%)               \\ \hline
\multicolumn{1}{|l|}{\textbf{Language}}                   & English vs. Other (\% of Tasks Favoring English)   & \textbf{1 (100\%)}           & 14 (0\%)        & 3 (100\%)               \\ \hline
\multicolumn{1}{|l|}{\multirow{5}{*}{\textbf{Ethnicity}}} & White vs. Other (\% of Tasks Favoring White)       & \textbf{0 (0\%)}         & 17 (76\%)        & 9 (22\%)                \\
\multicolumn{1}{|l|}{}                           & Black vs. Other (\% of Tasks Favoring Black)       & \textbf{2 (0\%)}           & 14 (71\%)            & 9 (22\%)               \\
\multicolumn{1}{|l|}{}                           & Hispanic vs. Other (\% of Tasks Favoring Hispanic) & \textbf{7 (0\%)}             & 15 (0\%)             & 16 (100\%)              \\
\multicolumn{1}{|l|}{}                           & Asian vs. Other (\% of Tasks Favoring Asian)       & \textbf{6 (50\%)}           & 1 (100\%)           & 6 (50\%)               \\
\multicolumn{1}{|l|}{}                           & "Other" vs. Other (\% of Tasks Favoring "Other")   & \textbf{10 (0\%)}             & 2 (0\%)           & 5 (100\%)              \\ \hline
\multicolumn{1}{|l|}{\multirow{3}{*}{\textbf{Insurance}}} & Medicare vs. Other (\% of Tasks Favoring Medicare) & \textbf{35 (83\%)}           & 51 (92\%)           & 48 (6\%)                \\
\multicolumn{1}{|l|}{}                           & Private vs. Other (\% of Tasks Favoring Private)   & \textbf{12 (8\%)}            & 41 (2\%)           & 44 (98\%)               \\
\multicolumn{1}{|l|}{}                           & Medicaid vs. Other (\% of Tasks Favoring Medicaid) & \textbf{19 (16\%)}           & 32 (19\%)           & 30 (83\%)               \\ \hline

\end{tabular}
\caption{Classifiers trained with baseline clinical BERT embeddings have multi-group fairness performance gaps (defined in Section \ref{sec:multigroup}) across gender, language, ethnicity, and insurance status. We count number of downstream classification tasks with a statistically significant differences (out of 57 total), as well as the percentage of significant tasks which \textit{favor} a subgroup. We correct for multiple hypotheses using the Benjamini-Hochberg procedure \cite{benjamini1995controlling}. Note that compared to Table \ref{tab:group_results}, most of the gaps still remain after controlling the false discovery rate.}
\label{tab:group_results_hyp} 
\end{table*}

\section{Adversarial Debiasing During Pretraining}
\subsection{Method}
\label{sec:appendix_adversial}
We extend existing methods for adversarial debiasing to BERT training. First, encoded sequences $x_1$ and $x_2$ are fed into BERT to yield the representation $h=f(x_1, x_2)$. In addition to information captured by the last-layer representation of the \texttt{[CLS]} token for the next-sentence task, we simultaneously feed $h_{[CLS]}$ to $a_1$ and $a_2$. $a_1$ is a discriminator minimizes the loss between the true protected group variable for $x_1$ (denoted as $z_1$) and $\hat{z}_1=a_1(h)$. Similarly, $a_2$ similarly tries to recover $z_2$ for $x_2$. The final loss function is:
\begin{equation}
    L=\sum_{(x_1, x_2)\in X} L_{adv}(a_1(J(h)), z_1) + L_{adv}(a_2(J(h)), z_2) + L_{LM} + L_{NS}
\end{equation}

Where $L_{LM}$ and $L_{NS}$ are the losses associated with the masked word prediction and next sentence prediction respectively. $J$ is an identity function with a negative gradient: $J(h)=h$, $\frac{dJ}{dx_1} = -\lambda \frac{dh}{dx_1}$. $\lambda$ is a hyperparameter which balances the utility to fairness trade-off. We use $\lambda =1$ in this work.

\subsection{Gender Debiasing}
\begin{table*}[hb!]
\begin{tabular}{l|l|l}
                   & \textbf{Baseline} & \textbf{Debiased} \\ \hline
\textbf{AUROC}     & 0.9168            & 0.8668            \\
\textbf{Precision} & 0.8480            & 0.8425            \\
\textbf{Recall}    & 0.7524            & 0.6242            \\
\textbf{AUPRC}     & 0.8954            & 0.8427            \\
\textbf{Log Loss}  & 0.3966            & 0.4724           
\end{tabular}
\caption{\label{tab:predict_groups} Comparison between the ability of baseline BERT versus debiased BERT representations to predict gender, using a fully connected neural network. }

\end{table*}

\clearpage
\subsection{Other Protected Groups}
\label{subsection:other_groups}

\begin{table}[h]
\begin{tabular}{|c|c|c|c|c|c|c|}
\hline
\multicolumn{7}{|c|}{\textbf{Significant Gap Count (\% Favouring English)}}                                                                                                      \\ \hline
\multirow{2}{*}{\textbf{Model}} & \multicolumn{3}{c|}{\textbf{Without FDR Correction}}                 & \multicolumn{3}{c|}{\textbf{With FDR Correction}}                    \\ \cline{2-7} 
                                & \textbf{Recall Gap} & \textbf{Parity Gap} & \textbf{Specificity Gap} & \textbf{Recall Gap} & \textbf{Parity Gap} & \textbf{Specificity Gap} \\ \hline
Baseline                        & \textbf{7 (29\%)}   & 17 (12\%)           & 9 (89\%)                 & \textbf{1 (100\%)}  & 14 (0\%)            & 3 (100\%)                \\ \hline
Debiased                        & \textbf{7 (43\%)}   & 20 (20\%)           & 11 (73\%)                & \textbf{5 (60\%)}   & 13 (23\%)           & 0 (0\%)                  \\ \hline
\end{tabular}
\caption{Fairness comparison of downstream performance using baseline and debiased embeddings, stratified by English vs Non-English speakers. The percentage of gaps favouring the English group is reported in the bracket. We collapse all other language groups for this comparison, due to small individual group sizes. The Benjamin-Hochberg procedure is used for false discovery rate correction~\cite{benjamini1995controlling}. Three different definitions of fairness are assessed.}
\end{table}


\begin{table}[]
\begin{tabular}{|c|c|c|c|c|c|}
\hline
\multirow{2}{*}{\textbf{Gap}}             & \multirow{2}{*}{\textbf{Group}} & \multicolumn{2}{c|}{\textbf{With FDR Correction}} & \multicolumn{2}{c|}{\textbf{Without FDR Correction}} \\ \cline{3-6} 
                                          &                                 & \textbf{Baseline}       & \textbf{Debiased}       & \textbf{Baseline}         & \textbf{Debiased}        \\ \hline
\multirow{5}{*}{\textbf{Recall Gap}}      & \textbf{Asian}                  & 8 (62\%)                & 7 (29\%)                & 6 (50\%)                  & 7 (29\%)                 \\ \cline{2-6} 
                                          & \textbf{Black}                  & 5 (20\%)                & 7 (71\%)                & 2 (0\%)                   & 4 (50\%)                 \\ \cline{2-6} 
                                          & \textbf{Hispanic/Latino}        & 7 (0\%)                 & 6 (0\%)                 & 7 (0\%)                   & 6 (0\%)                  \\ \cline{2-6} 
                                          & \textbf{White}                  & 4 (75\%)                & 6 (100\%)               & 0 (0\%)                   & 1 (100\%)                \\ \cline{2-6} 
                                          & \textbf{Other}                  & 10 (0\%)                & 11 (9\%)                & 10 (0\%)                  & 10 (0\%)                 \\ \hline
\multirow{5}{*}{\textbf{Parity Gap}}      & \textbf{Asian}                  & 7 (100\%)               & 4 (100\%)               & 1 (100\%)                 & 1 (100\%)                \\ \cline{2-6} 
                                          & \textbf{Black}                  & 18 (72\%)               & 24 (71\%)               & 14 (71\%)                 & 24 (71\%)                \\ \cline{2-6} 
                                          & \textbf{Hispanic/Latino}        & 18 (0\%)                & 17 (6\%)                & 15 (0\%)                  & 11 (9\%)                 \\ \cline{2-6} 
                                          & \textbf{White}                  & 22 (82\%)               & 20 (85\%)               & 17 (76\%)                 & 16 (81\%)                \\ \cline{2-6} 
                                          & \textbf{Other}                  & 8 (0\%)                 & 6 (0\%)                 & 2 (0\%)                   & 7 (100\%)                \\ \hline
\multirow{5}{*}{\textbf{Specificity Gap}} & \textbf{Asian}                  & 8 (62\%)                & 7 (29\%)                & 6 (50\%)                  & 6 (50\%)                 \\ \cline{2-6} 
                                          & \textbf{Black}                  & 5 (20\%)                & 7 (71\%)                & 9 (22\%)                  & 6 (17\%)                 \\ \cline{2-6} 
                                          & \textbf{Hispanic/Latino}        & 7 (0\%)                 & 7 (71\%)                & 16 (100\%)                & 7 (100\%)                \\ \cline{2-6} 
                                          & \textbf{White}                  & 4 (75\%)                & 6 (100\%)               & 9 (22\%)                  & 5 (20\%)                 \\ \cline{2-6} 
                                          & \textbf{Other}                  & 10 (0\%)                & 11 (9\%)                & 5 (100\%)                 & 7 (100\%)                \\ \hline
\end{tabular}
\caption{Fairness comparison of downstream performance using baseline and debiased embeddings, stratified by ethnicities. The number of tasks with a statistically significant gap is reported, with the percentage of tasks favouring that subgroup presented in brackets. The Benjamin-Hochberg procedure is used for false discovery rate correction~\cite{benjamini1995controlling}. Three different definitions of fairness are assessed.}
\end{table}

\begin{table}[]
\begin{tabular}{|c|c|c|c|c|c|}
\hline
\multirow{2}{*}{\textbf{Gap}}             & \multirow{2}{*}{\textbf{Group}} & \multicolumn{2}{c|}{\textbf{With FDR Correction}} & \multicolumn{2}{c|}{\textbf{Without FDR Correction}} \\ \cline{3-6} 
                                          &                                 & \textbf{Baseline}       & \textbf{Debiased}       & \textbf{Baseline}         & \textbf{Debiased}        \\ \hline
\multirow{3}{*}{\textbf{Recall Gap}}      & \textbf{Medicaid}               & 20 (20\%)               & 16 (0\%)                & 19 (16\%)                 & 12 (0\%)                 \\ \cline{2-6} 
                                          & \textbf{Medicare}               & 33 (85\%)               & 34 (88\%)               & 35 (83\%)                 & 35 (86\%)                \\ \cline{2-6} 
                                          & \textbf{Private}                & 15 (7\%)                & 14 (14\%)               & 12 (8\%)                  & 12 (8\%)                 \\ \hline
\multirow{3}{*}{\textbf{Parity Gap}}      & \textbf{Medicaid}               & 31 (19\%)               & 32 (12\%)               & 32 (19\%)                 & 34 (12\%)                \\ \cline{2-6} 
                                          & \textbf{Medicare}               & 51 (92\%)               & 49 (94\%)               & 51 (92\%)                 & 52 (94\%)                \\ \cline{2-6} 
                                          & \textbf{Private}                & 41 (2\%)                & 41 (2\%)                & 41 (2\%)                  & 44 (5\%)                 \\ \hline
\multirow{3}{*}{\textbf{Specificity Gap}} & \textbf{Medicaid}               & 30 (83\%)               & 29 (83\%)               & 30 (83\%)                 & 29 (83\%)                \\ \cline{2-6} 
                                          & \textbf{Medicare}               & 48 (6\%)                & 46 (2\%)                & 48 (6\%)                  & 48 (4\%)                 \\ \cline{2-6} 
                                          & \textbf{Private}                & 40 (98\%)               & 41 (95\%)               & 44 (98\%)                 & 42 (95\%)                \\ \hline
\end{tabular}
\caption{Fairness comparison of downstream performance using baseline and debiased embeddings, stratified by insurance. The number of tasks with a statistically significant gap is reported, with the percentage of tasks favouring that subgroup presented in brackets. The Benjamin-Hochberg procedure is used for false discovery rate correction~\cite{benjamini1995controlling}. Three different definitions of fairness are assessed.}
\end{table}

\end{document}